\documentclass[conference]{IEEEtran}
\IEEEoverridecommandlockouts
\usepackage{amsmath}
\usepackage{amssymb}
\usepackage{xspace}
\usepackage{microtype}
\usepackage{color}
\usepackage{url}
\usepackage{lipsum}
\usepackage{fancyhdr}
\usepackage{array}
\usepackage{comment}

\usepackage{cite}
\usepackage{graphicx}
\usepackage{textcomp}
\usepackage{mathtools}
\usepackage{amsmath,mleftright}
\usepackage{algpseudocode}
\usepackage{xparse}
\usepackage{comment}
\usepackage{amssymb}
\usepackage{mathbbol}
\usepackage{amsthm}
\usepackage{hyperref}
\usepackage{graphicx}% http://ctan.org/pkg/graphicx
\usepackage{amsmath}
\usepackage{amssymb}
\usepackage{enumitem}
\usepackage{cite}
\usepackage{tabularx}
\usepackage[table]{xcolor}
\usepackage{subfig}
\usepackage[ruled,vlined,noend]{algorithm2e}
\usepackage{float}
\def\BibTeX{{\rm B\kern-.05em{\sc i\kern-.025em b}\kern-.08em
    T\kern-.1667em\lower.7ex\hbox{E}\kern-.125emX}}
    
\begin{document}

\title{SLAM-based Safe Indoor Exploration Strategy
\thanks{Corresponding Author email: anthony.tzes@nyu.edu}
}

\author{
\IEEEauthorblockN{Omar Mostafa}
\IEEEauthorblockA{\textit{Center for Artificial Intelligence \& Robotics (CAIR)} \\
\textit{New York University Abu Dhabi (NYUAD)}\\
United Arab Emirates \\
omm7813@nyu.edu}
~\\
\and
\IEEEauthorblockN{Nikolaos Evangeliou}
\IEEEauthorblockA{\textit{Robotics \& Intelligent Systems Control Lab} \\
\textit{NYUAD}\\
United Arab Emirates \\
nikolaos.evangeliou@nyu.edu}
~\\
\and
\IEEEauthorblockN{Anthony Tzes\textsuperscript{*}}
\IEEEauthorblockA{\textit{CAIR} \\
\textit{NYUAD}\\
United Arab Emirates \\
anthony.tzes@nyu.edu}
~\\
}

\maketitle
\begin{abstract} 
This paper suggests a 2D exploration strategy for a planar space cluttered with obstacles. Rather than using point robots capable of adjusting their position and altitude instantly, this research is tailored to classical agents with circular footprints that cannot control instantly their pose. Inhere, a self-balanced dual-wheeled differential drive system is used to explore the place. The system is equipped with linear accelerometers and angular gyroscopes, a 3D-LiDAR, and a forward-facing RGB-D camera. The system performs RTAB-SLAM using the IMU and the LiDAR, while the camera is used for loop closures. The mobile agent explores the planar space using a safe skeleton approach that places the agent as far as possible from the static obstacles. During the exploration strategy, the heading is towards any offered openings of the space. This space exploration strategy has as its highest priority the agent's safety in avoiding the obstacles followed by the exploration of undetected space. Experimental studies with a ROS-enabled mobile agent are presented indicating the path planning strategy while exploring the space.
\end{abstract} 

\begin{IEEEkeywords}
Space Exploration, RTAB-SLAM, Safety, Skeleton-based Path Planning
\end{IEEEkeywords}

\section{Introduction}

%On SLAM and ICP-Odometry
The usage of LiDAR and cameras in SLAM (Simultaneous Localization and Mapping) is popular in the area of visual-SLAM and LiDAR-SLAM~\cite{lluvia2021active}. In such concept, the distance measurements offered by LiDAR, using Time-of-Flight (ToF) methods, can be complemented by visual inputs from depth camera(s) to ensure robust mapping and localization using loop closures~\cite{ismail2022exploration}. RTAB-Map, an open-source SLAM library, is employed to perform SLAM by integrating both camera and LiDAR data. RTAB-Map has evolved from an appearance-based loop closure detection system into a comprehensive SLAM solution, capable of managing large-scale and long-term operations~\cite{labbe2019rtab}. Accurate localization can be achieved by continuously matching recent sensor data with an occupancy grid map to reduce the cumulative errors encountered in dead reckoning methods~\cite{yamauchi1998mobile}. The proposed system leverages the continuous localization provided by RTAB-Map, which frequently updates the robot’s location based on sensor data with respect to the map frame. This method significantly reduces positional errors over time and enhances the overall accuracy and reliability of the SLAM process.

%On Path Planning and Navigation
In obstacle-rich environments, safe path planning and navigation are essential in the exploration strategy. To achieve safe trajectories, visibility and connectivity graphs represent the environment with high complexity and hence are computationally inefficient. This complexity can be reduced by achieving its minimal skeleton representation~\cite{lam1992thinning}. This technique provides a minimal skeleton by iteratively removing pixels from the environment's representation. The abstained skeleton represents the core pathways that the robot can navigate through~\cite{beom1993path}. Similarly, the image-skeleton can be computed using the Shortest Feasible Path (SFP) accounting for nonholonomic constraints. The SFP-metric captures the difficulty of moving between configurations, enabling the robot to navigate efficiently~\cite{mirtich1992using}. %Moreover, the skeleton-extraction can be combined with a greedy algorithm to reduce time consumption and ensure that the robot maintains a safe distance from the obstacles throughout the path. The greedy algorithm optimizes the path by directly connecting new nodes to existing nodes in the search tree without encountering obstacles~\cite{chang2022skeleton}. 

The Navigation2 software~\cite{macenski2023survey} is used for navigation, enhanced with the proposed skeleton-based path-planning algorithm. This ensures safe path planning, while navigating through goal points during the exploration process.

%On Exploration Strategy
Exploration algorithms typically rely on frontier-based exploration, where robots target boundary regions between known and unknown spaces \cite{keidar2014efficient,yamauchi1998mobile,luperto2020robot}. This approach prioritizes accessible areas and expands the explored environment. Another approach for goal selection is information-based, which moves the robot to information rich locations, according to some information measures~\cite{luperto2020robot}. %In~\cite{lluvia2021active}, safe regions (obstacle-free areas) were proposed for map-creation and path-planning. 
The goal positions are decided during exploration and depend on the Next Best View algorithm~\cite{lauri2020multi}, as the ones that would most likely explore a large area of the environment. 

The proposed exploration algorithm is designed to handle practical exploration tasks, tolerating noisy sensor measurements and maintaining accuracy in dynamic indoor settings. The LiDAR PointCloud provides all available headings for the robot’s current position and uses the ones that guarantee new observed areas. By focusing on safe navigation and forward-directed exploration, the proposed method prioritizes obstacle avoidance while effectively expanding the explored space.

Experimental validation with a self-balanced, dual-wheeled robot in an obstacle-rich real-world apartment demonstrates the robustness of the approach. The results showcase the algorithm's ability to mitigate sensor noise, avoid drifts, and adapt to challenging exploration scenarios, making it a reliable solution for practical applications.

% Our exploration strategy uses safe regions in the robot’s forward-facing semi-planes. The LiDAR PointCloud provides all available headings for the robot’s current position and uses the ones that guarantee new observed areas. The devised algorithm restricts the robot’s waypoints to be inside the polygon constructed by the intersection of the forward semi-planes. 
%
\section{Problem Statement}
The  RTAB-Map provides 2D and 3D Occupancy grids~\cite{labbe2019rtab}. The IMU-sensors (angular velocity and linear acceleration) were fed to RTAB-Map centered at the body of the self-balanced Diablo-robot~\cite{liu2024diablo6dofwheeledbipedal}. The 3D-LiDAR offers high-resolution point cloud data used for a detailed occupancy grid that serves as the foundation of the robot’s internal map. At the same time, the RGB-D camera provides visual odometry data, enabling the system to address cumulative localization errors, "drifts," through visual loop closures~\cite{labbe2019rtab}. It recognizes previously visited locations and uses them as benchmarks to recalibrate the robot's position and correct the map. This feature is crucial in indoor experiments (such as a home apartment) which have repetitive and visually similar spaces. In such an environment, the ability of RTAB-Map to detect loop closures and correct for drift is crucial for maintaining an accurate map and ensuring reliable navigation throughout the space.

RTAB-Map parameters were tuned to provide minimum drifts, while the employed graph optimization strategy is General Graph Optimization (g2o)~\cite{kummerle2011g}. Ten iterations were used in the optimizer for computing the solution of the graph optimization; this reduced the impact of drift. The system's odometry relies on the Iterative Closest Point~\cite{zhang2021fast} where the employed SLAM was limited to 3 DoFs ($xy$-position and yaw). The LiDAR measurements were limited to 5m resulting in a low complexity Point Cloud. 
\section{Space Exploration and Safe Path Planning}
\subsubsection*{Mathematical Notation} \leavevmode
Let at robot waypoint $k$, the robot's center of mass position~(orientation) be $p_i^k=\left[x_i^{k,r},y_i^{k,r}\right]^{\top}~(r_i^{k,z})$, where $x_i^{k,r} \land y_i^{k,r} >0$, $r_i^{k,z} \in [0^{\circ},360^{\circ})$ and $i=1,\ldots,N, N \in (0,\infty)$ the number of potential directions the robot can move to. We should note that this allows variable time periods between waypoints and is not a reflection of the time instant. The half-plane at the $k$th waypoint that points forward to the vehicle is ${\cal H}^{k}(P_i,r_i^z):y - \tan \left(90^{\circ} + r_i^{k,z}\right) x -\left[ y_i^r - \tan\left(90^{\circ} + r_i^{k,z}\right) x_i^r \right] \geq 0$. 
Let the intersection over $L$-samples of all half planes be ${\cal H}^{k}_{k-L-1}=\cap_{i=k-L+1}^{k} {\cal H}^{i}$.

\begin{figure} [htbp]
  \centering
    \includegraphics[width=0.7\linewidth]{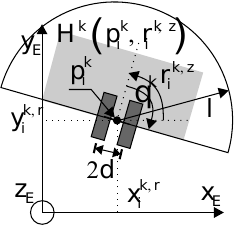}
    \caption{Robot top view pose mathematical notation.}
    \label{fig:robo_math}
\end{figure}

Assume 
%the vehicle's 2-wheels be separated by $2d$; the right-wheel point which connects the differential is $p_w^{k,r} = p_i^k + d \left[ \cos(r_i^{k,z}),-\sin(r_i^{k,z}) \right]^{\top}$. It is assumed 
that a forward aligned LiDar sensor is mounted at the center of the robot and can detect the presence of obstacles within a distance $l$ and a semicircle $\theta_l^k \in \left[ r_i^{k,z}-90^{\circ}, r_i^{k,z}+90^{\circ} \right]$, as shown in Figure~\ref{fig:robo_math}.

\subsection{Safe Exploration Strategy}
\begin{figure} [htbp]
  \centering
    \includegraphics[width=0.7\linewidth]{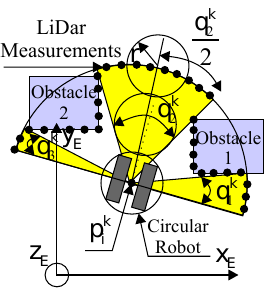}
    \caption{Forward allowable regions $q_i^k,~i=1,2,3$ and $L=1$.}
    \label{fig:mintheta}
\end{figure}

The forward scanning LiDar scans in ${\cal H}_{k-L+1}^k$. If several obstacles are detected (like Obstacle 1 and 2 in Figure~\ref{fig:mintheta}), the safe exploration component computes the potential directions that the robot can move to $q^k \in \left\{ q_1^k,\ldots,q_n^k \right\}, \land q_i^k\in \left[q_i^{k,-},q_i^{k,+}\right]$; where in Figure~\ref{fig:mintheta}, $n=3$ and $L=1$.

The safety component computes the optimal heading by selecting the largest region that: a) can be explored and b) is farthest away from obstacles. This region corresponds to $q_{i^{k,*}}=\min_{i=1}^n q_i^k$ and the robot moves towards the skeleton (dichotomous angle of this region). In Figure~\ref{fig:mintheta} this corresponds to $i^{k,*}=2~(q_2^k)$ while the optimal heading is $\frac{q_2^k}{2}$ computed in the available angle frame. The inherent assumption is that the robot can pass through the obstacles while this heading is selected. For rotation invariance purposes the robot is modeled as circular with radius $r$; in Figure~\ref{fig:mintheta} the three circles display the consequent motion of the robot% towards its subsequent waypoint
, while the current LiDar measurements are indicated with black dots.
\subsection{Waypoint Selection}
\begin{figure} [htbp]
  \centering
    \includegraphics[width=0.7\linewidth]{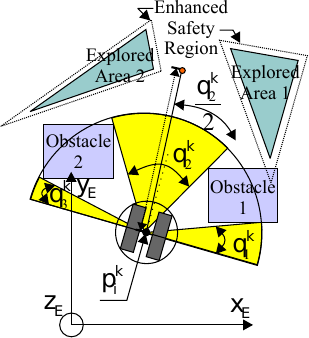}
    \caption{Subsequent waypoint selection}
    \label{fig:next_waypoint}
\end{figure}

Having identified the safe heading of the robot motion, the subsequent waypoint is computed along this heading at a distance $\tilde{d}$. To emphasize the simplicity of the algorithm this distance is kept constant $(\tilde{d} \geq l)$ throughout the motion, as shown in Figure~\ref{fig:next_waypoint} for two explored areas. If this point has not been visited before, then it becomes the new $(k+1)$-waypoint and the robot moves according to
\begin{eqnarray}
    x_{i+1}^{k+1,r} & = & x_i^{k,r} + \tilde{d} \cos \left( q_{i^{*}}^{k,-} + \frac{ q_{i^{k,*}}}{2} \right) \\
    y_{i+1}^{k+1,r} & = & y_i^{k,r} + \tilde{d} \sin \left( q_{i^{*}}^{k,-} + \frac{ q_{i^{k,*}}}{2} \right) \\
    r_{i+1}^{k+1,z} & = & q_{i^{k,*}}^{-} + \frac{ q_{i^{k,*}}}{2}
\end{eqnarray}
Indirectly this implies that the vehicle will move to its new waypoint with the ``safe orientation'' $r_{i+1}^{k+1,z}$. For every waypoint the reference pose $\left[ x_{i+1}^{k+1,r},y_{i+1}^{k+1,r},r_{i+1}^{k+1,r}\right]^{\top}$of the robot should be recorded. If this point has been revisited, then the other available safe angles should be tried, or $i^{k,*} \in \left\{ 1,\ldots,n \right\} - i^{k,*}$. While moving from waypoint $p_i^k$ to $p_i^{k+1}$ the robot should pass between obstacles (see for example Obstacle \#1 and \#2 in Figure~\ref{fig:next_waypoint}). However traditional methods~\cite{urakubo2001motion,klancar2005mobile,arvanitakis2012trajectory} cannot guarantee a straight trajectory. In this case, the navigation used in \cite{macenski2023survey} is employed. In this package, the robot's kinodynamic model is 
\begin{eqnarray}
    \left[ 
    \begin{array}{c}
    \dot{p}_i^k \\ \hline
    \dot{r}_i^{k,z}
    \end{array}
    \right]
    &=&
    \left[
    \begin{array}{c|c}
    \cos\left(r_i^{k,z}\right) & 0\\
    \sin\left(r_i^{k,z}\right) & 0 \\ \hline
    0 & 1
    \end{array}
    \right]
    \left[ \begin{array}{c} v_i^k \\ \hline \omega_i^{k,z}\end{array} \right],~\mbox{where}\\
    v_i^{k} & = & \sqrt{\dot{p}_i^k},~
    \omega_i^{k,z} = \frac{\dot{x}_i^{k,r} \ddot{y}_i^{k,r} - \dot{y}_i^{k,r} \ddot{x}_i^{k,r}}{\left(p_i^{k}\right)^2}.
\end{eqnarray}

{\bf Lemma 1:}
    In an obstacle-free environment, the exploration strategy moves the robot forward at a constant orientation equal to the initial one.\newline
{\em Proof:} In the absence of obstacles, $n=1$ and $q_i^{k,+}=-q_1^{k,+}=90^{\circ}$ while the robot moves along the initial orientation $r_i^{1,z}$. \qed

\begin{figure} [htbp]
  \centering
    \includegraphics[width=0.85\linewidth]{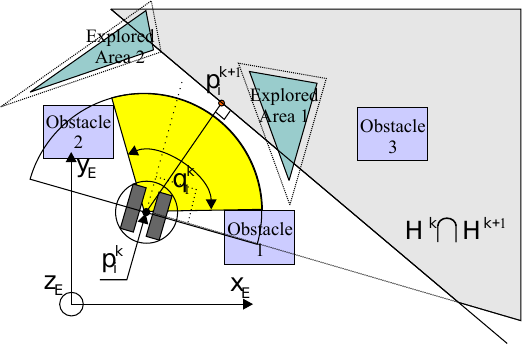}
    \caption{Feasible Exploration Area}
    \label{fig:next_halfplane}
\end{figure}

For the case where $L=1$, the robot at the $k+1$-waypoint, rather than moving in the halfplane ${\cal H}^{k+1}$, it moves in the area ${\cal H}^k \cap {\cal H}^{k+1}$, as shown in Figure~\ref{fig:next_halfplane}. This selection constrains the halfplane ${\cal H}^{k+1}$ into a smaller region, yet restricts the robot's motion to a forward one while exploring the space.

\begin{figure} [htbp]
  \centering
    \includegraphics[width=0.85\linewidth]{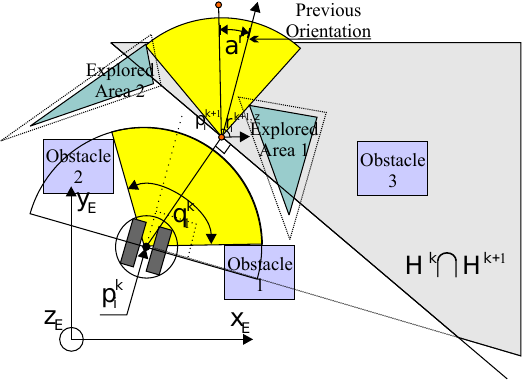}
    \caption{Neighboring Waypoint Orientation Case}
    \label{fig:next_orientation}
\end{figure}

For every waypoint that may have been visited before, the exploration strategy needs to keep the complete pose (position and orientation) of the vehicle. Each waypoint may belong to a previously explored area depending on whether the underlying SLAM algorithm has deemed it as such. In order to avoid moving in the neighborhood to these points, if there is no considerable difference $\parallel r_{i+1}^{k+1,z} \cdot r_{i+2}^{k+2} \parallel \leq a^r$, as shown in Figure~\ref{fig:next_orientation}, the system will use the next potential angles (if any) and will continue.

\begin{figure} [htbp]
  \centering
    \includegraphics[width=0.85\linewidth]{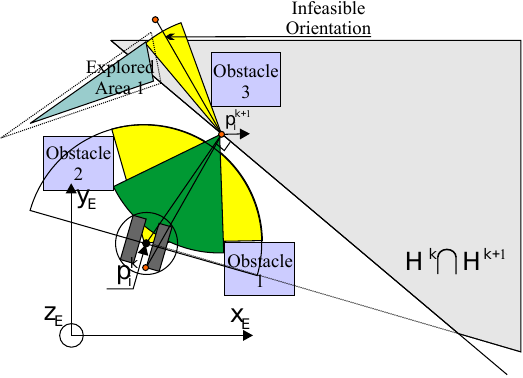
    }
    \caption{Reverse Navigation Case}
    \label{fig:reverse}
\end{figure}

For the case where ${\cal H}^k_{k=L+1} = \emptyset$, or, when $p_i^{k+1} \in$ Explored Area \# or any Obstacle \#, then the forward motion stops and needs to be reversed, according to Figure~\ref{fig:reverse}. In this case, the rule of moving in previous neighboring sites is not activated by setting $a^r=1$. The reversed feasible area is shown in Figure~\ref{fig:reverse} with a "Grass Green" color

The detailed steps of the exploration strategy are presented in the pseudo-code [\ref{alg:exploration}]. Note that $stop\_exploration$ will remain $False$ unless changed by the user.

\begin{algorithm}[h!]
\caption{Proposed Area Exploration Algorithm} \label{alg:exploration}
\KwIn{$\mathcal{H}^0$, $l$, $a^r$, $\{P_i \mid i = 0, \dots, N\}$, $q_n^k = \left\{ q_1^k,\ldots,q_n^k \right\}$ (in descending order of safety)}

Set flag $stop\_exploration \gets \text{False}$\;
Set $k \gets 0$\;

Define \texttt{FindNextHeading}($\mathcal{H}_1^k$, $q_n^k$) as:\\
    \ForEach{potential heading $q^k \in q_n^k$}{
        Calculate the goal position $p_i^{k+1}$.\\
        
        \If{$ \forall p \in P_{i}, \; \| p_i^{k+1} - p \| \geq l$}{
            Set $q_{i^{k,*}} \gets q^k$\;
            \textbf{break}\;
        }
        \ElseIf{$ p \in P_{i}, \; \| p_i^{k+1} - p \| \leq l$}{
            \If{$\| r_{i+1}^{k+1,z} \cdot r_{i+2}^{k+2} \| \leq a^r$}{
                Set $q_{i^{k,*}} \gets q^k$\;
                \textbf{break}\;
            }
        }
    }
return $q_{i^{k,*}}$

\While{$stop\_exploration = \text{False}$}{
    Find current forward half-plane $\mathcal{H}^k$\;
    Compute $\mathcal{H}_{k-L-1}^k$ such that $L = k$\;

    Set $q_{i^{k,*}} \gets$ \texttt{FindNextHeading}($\mathcal{H}_1^k$, $q_n^k$)\;
    
    \If{$q_{i^{k,*}} \neq NULL$}{
        Move to $p_i^{k+1}$\;
    }
    \ElseIf{$q_{i^{k,*}} = NULL$}{
        Set $\mathcal{H}_1^k \gets \emptyset$\;
        \texttt{FindNextHeading}($\mathcal{H}_1^k$, $q_n^k$)\;
        
        \If{$q_{i^{k,*}} = NULL$}{
            Set $q_{i^{k,*}} \gets q_{i^{k,*}} + 180^\circ \mod 360^\circ$\;
        }
    }
}
\end{algorithm}
\section{Robot SLAM}
The 2-wheeled robot is equipped with a $360^{\circ}$ LiDar sensor, a forward-looking RGB-D sensor and a triple Inertial Measurement Unit. Wheel odometry is not used in Robot Simultaneous Localization And Mapping (SLAM) due to the noisy measurements. SLAM enhances the effectiveness of the explored map by allowing large coverage \cite{tzes2018visual} problems, while the understanding of the environment's map and the robot's location within it is its cornerstone. Single robot SLAM for both 2D and 3D motion is a mature field with many advanced platforms and frameworks~\cite{kazerouni2022survey, huang2023indoor}. The SLAM task requires the use of dense, rich data from 3D data streams to generate a useful and accurate environment representation, thus 
%Real-Time Appearance-Based Mapping (RTAB-MAP)~\cite{labbe2019rtab} is an open-source library implementing loop closure detection under a fixed time limit. RTAB-Map has been extended to a complete graph-based SLAM approach.
% and unlike techniques which use only LiDar~\cite{dellenbach2022ct}, it uses its camera for loop closures.
the 3D occupancy from RTAB-Map, represented as octree voxels, is incorporated. The latter are projected to the ground to create 2D maps using pixels~\cite{unlu2023path_planner}. A region can be classified as untraversable, or its opposite. A frontier is any point on the contour of a traversable region. The path planning algorithm follows that of~\cite{unlu2023path_planner} enhanced by the aforementioned steps of the space exploration.
\section{Experimental Studies}
\subsection{Mobile Agent Prototype}
\begin{figure} [htbp]
    \centering
    \includegraphics[width=0.7\linewidth]{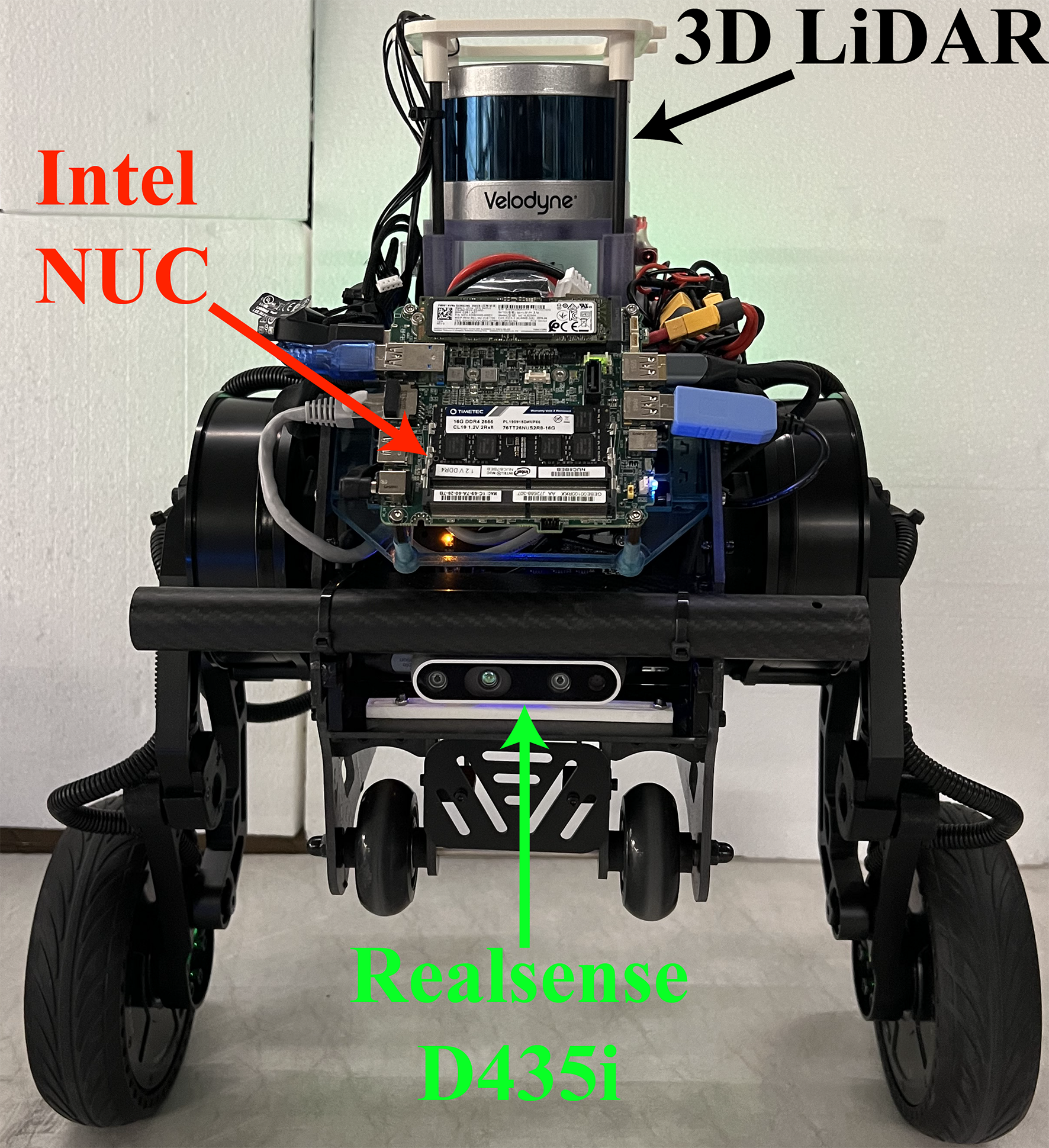}
    \caption{Diablo two-wheeled self-balanced robot.}
    \label{fig:Diablo_robot}
\end{figure}

The self-balanced two-wheeled Diablo robot agent shown in Figure~\ref{fig:Diablo_robot} was utilized%~\cite{liu2024diablo6dofwheeledbipedal}
. %This platform has agility to pass through the corridors encountered in indoor household environments. 
The agent was fitted with onboard sensory equipment including: a) a Velodyne VLP-16 LiDAR, b) a forward-facing Intel Realsense D435i RGB-D camera and c)an Intel-i7 NUC computer for the computationally intensive tasks. The sensors and actuators communicate through the ROS middleware. The RTAB-SLAM was used by accounting for the LiDAR, the 2X linear accelerometers, and the 2X gyroscopes. The $z$-altitude was constrained to $z=0$ since the mobile agent was moving at a planar surface parallel to the ground.

\subsection{Exploration Strategy}
\begin{figure} [h!]
    \centering
    \includegraphics[width=0.75\linewidth]{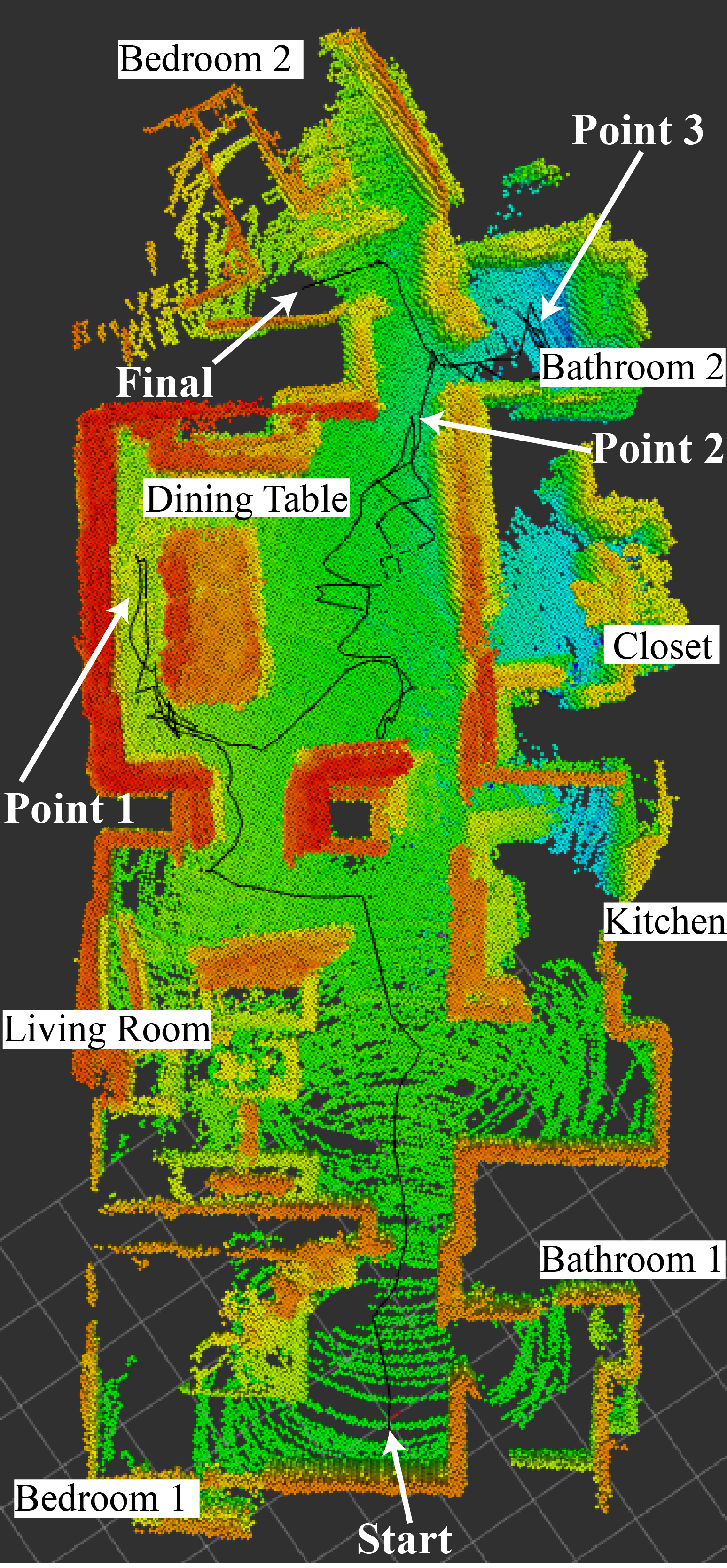}
    \caption{Robot SLAM and path planning during exploration}
    \label{fig:real_world_results}
\end{figure}

% \subsubsection{Experimental case} 
The proposed exploration algorithm was tested in the environment of an obstacle-cluttered apartment~\cite{huang2023fael}. This indoor area (apartment) had multiple rooms necessitating the continuous orientation adjustments of the robot which resulted in increased positional drifs. The derived map and the robot's path (black line) are displayed in Figure~\ref{fig:real_world_results} using the exploration strategy (after 3 minutes).

A detailed view of the exploration process highlights the efficiency of the proposed algorithm. At Point 1 (highlighted in Figure~\ref{fig:real_world_results}), the robot could not continue moving forward due to noise interference. Following the algorithm, it rotates to a new exploration backward direction and restarts the process; the parameter $\alpha$ prevents the robot from revisiting waypoints with the same orientation, but it allows backward movement. The same occurred at Point 2, but here the noise was removed by RTAB-Map after slight movement, allowing the robot to resume and visit Bathroom and Bedroom 2. At Point 3, the algorithm demonstrates its ability to guide the robot out of enclosed spaces, as it successfully exited Bathroom 2 after exploring the area. 

% \subsubsection{Simulation Case} To test the algorithm's validity (and robustness in sensor noise), a simulated apartment resembling the real-one was created in the Gazebo physics simulator. The Diablo-robot along with RTAB-Map and the Navigation2 stack were subsequently simulated. The resulting map along with the robot' trajectory is shown in Figure~~\ref{fig:simulation_results}, where the robot created an Octomap. Depending on the noise level the robot explored the place for 10 minutes. This led to the exploration of more spaces as if the robot didn't visit a location on the first pass due to the preferred safe headings; the robot would visit it on the return trip. This was exemplified in the experiment when the robot initially bypassed the kitchen on its first pass. However, after reaching and exploring Bedroom 2, the robot was able to retrace its path and subsequently visit the kitchen. %This demonstrates that the algorithm is more efficient in longer experiments, given an accurate map.
% \begin{figure} [htbp]
%     \centering
%     \includegraphics[width=0.8\linewidth]{figures/simulation_results(new_apart).pdf}
%     \caption{Generated Simulated Map and Robot's Path}
%     \label{fig:simulation_results}
% \end{figure}

\section{Conclusions}
This paper presented an approach for indoor exploration using a skeleton-based path planning algorithm integrated with RTAB-Map SLAM. By prioritizing safe headings and revisiting previously unexplored spaces, the algorithm ensures comprehensive coverage using noisy measurements. The experiments showed that the robot can reliably maintain its map accuracy while optimizing its path for safety in cluttered environments. 

Future work will focus on extending this exploration framework to multi-robot systems for collaborative exploration of unknown areas. The goal is to enable multiple robots to efficiently divide and explore the environment while dynamically sharing information to construct a unified map. 

\bibliographystyle{ieeetr}  
\bibliography{references}

\end{document}